\pdfoutput=1

\documentclass[11pt]{article}
\usepackage{pgfplotstable}
\usepackage{tikz}
\usepackage{pgfplots}
\pgfplotsset{compat=1.18}
\usepackage{booktabs}
\usepackage{siunitx}
\usepackage[final]{acl}

\usepackage{times}
\usepackage{tabularx}
\usepackage{latexsym}
\usepackage[T1]{fontenc}
\usepackage[utf8]{inputenc}
\usepackage{microtype}
\usepackage{inconsolata}
\usepackage{graphicx}
\usepackage{multirow}
\usepackage{array}
\usepackage{amsmath}  
\usepackage{xcolor}
\usepackage{tcolorbox}
\newenvironment{promptbox}{\begin{tcolorbox}[colback=gray!10,colframe=black!80]}{\end{tcolorbox}}
\usepackage{pgf-pie}

\title{Agent-in-the-Loop: A Data Flywheel for Continuous Improvement in LLM-based Customer Support}

\usepackage[final]{acl}

\title{Agent-in-the-Loop: A Data Flywheel for Continuous Improvement in LLM-based Customer Support}

\usepackage[final]{acl}

\title{Agent-in-the-Loop: A Data Flywheel for Continuous Improvement in LLM-based Customer Support}

\author{
  \textbf{Cen (Mia) Zhao} \quad \textbf{Tiantian Zhang} \quad \textbf{Hanchen Su} \quad \textbf{Yufeng (Wayne) Zhang} \\
  \textbf{Shaowei Su} \quad \textbf{Mingzhi Xu} \quad \textbf{Yu (Elaine) Liu} \quad \textbf{Wei Han} \\
  \textbf{Jeremy Werner} \quad \textbf{Claire Na Cheng} \quad \textbf{Yashar Mehdad} \\
  Airbnb, Inc., USA \\
  \texttt{\{mia.zhao, tiantian.zhang, hanchen.su, wayne.zhang, shaowei.su, mingzhi.xu,} \\
  \texttt{elaine.liu, wei.han, jeremy.werner, claire.cheng, yashar.mehdad\}@airbnb.com}
}

\begin{document}
\maketitle
\begin{abstract}
We introduce an \emph{Agent-in-the-Loop} (AITL) framework that implements a continuous \emph{data flywheel} for iteratively improving an LLM-based customer support system. Unlike standard offline approaches that rely on batch annotations, AITL integrates four key types of annotations directly into live customer operations: (1) pairwise response preferences, (2) agent adoption decisions and rationales, (3) knowledge relevance checks, and (4) identification of missing knowledge. These feedback signals seamlessly feed back into model updates, reducing retraining cycles from months to weeks. Our production pilot involving US-based customer support agents demonstrated significant improvements in retrieval accuracy (+11.7\% recall@75, +14.8\% precision@8), generation quality (+8.4\% helpfulness), and agent adoption rates (+4.5\%). These results underscore the effectiveness of embedding human feedback loops directly into operational workflows to continuously refine LLM-based customer support systems.
\end{abstract}

\section{Introduction}
\label{sec:intro} 
Retrieval-augmented generation (RAG) improves large language models (LLMs) by grounding responses to external knowledge, overcoming static limitations, and improving transparency through evidence-based outputs~\cite{Lewis2020RAG}. However, traditional LLMs, typically trained in a fixed dataset with static knowledge cut-off points, inherently struggle to adapt to evolving real-world interactions without interventions such as continuous learning or retrieval enhancement \cite{shah2023nlp}.

Recent research emphasizes the importance of a \emph{data flywheel}, an iterative feedback loop that continuously leverages new interaction data to enhance model performance~\citep{luo2024arenalearningbuilddata}. In customer support scenarios, such a data flywheel is particularly valuable due to evolving product features, shifting user preferences, and continuously updated policies and procedures.
~\citet{dai2025are}'s daily oracle benchmark demonstrates that static models, even when paired with retrieval, lose more than 20 percentage points of accuracy on news questions within a few years, indicating that continuous feedback loops are crucial for preventing drift and maintaining relevance in real-world systems. 

\paragraph{Our Contributions.}
To maintain a continuous human-driven data flywheel for accurate and relevant customer support, we (1) develop an \emph{annotation interface} capturing response preferences, adoption rationales, knowledge relevance, and missing knowledge during live conversations, and (2) implement a \emph{continuous learning pipeline} that integrates these annotations into training datasets, reducing model update cycles from months to weeks. A US-based pilot confirms significant improvements in retrieval accuracy, response helpfulness, citation correctness, and agent adoption rates. To further optimize annotation efficiency at scale, we recommend delaying annotations for preference, adoption, and knowledge relevance, while immediately annotating missing knowledge when SLAs permit.

\section{Related Work}

To address critical issues such as preference drift and knowledge decay, recent research has integrated human or AI-simulated feedback within reinforcement learning frameworks.

\paragraph{Human-in-the-Loop and Preference Optimization.}
Human-in-the-loop (HITL) approaches enhance LLM alignment by directly optimizing outputs toward explicit human preferences, moving beyond traditional supervised learning metrics~\citep{stiennon2020learning}. Reinforcement Learning with Human Feedback (RLHF), pioneered by OpenAI~\citep{ouyang2022training}, aligns models with human preferences through pairwise comparisons obtained via offline annotations. Anthropic subsequently introduced iterative online human feedback loops, continuously incorporating real-time human annotations to significantly enhance conversational agent helpfulness and harmlessness~\citep{bai2022traininghelpfulharmlessassistant}. Further advancing alignment at scale, Anthropic proposed Constitutional AI, a method employing Reinforcement Learning from AI Feedback (RLAIF) guided by explicit human-defined principles~\citep{bai2022constitutionalaiharmlessnessai}.

\paragraph{Data Flywheel and Continuous Learning Pipelines.}
Recently, \citet{luo2024arenalearningbuilddata} introduced \emph{Arena Learning}, an automated data flywheel using simulated self-play between LLMs and AI judges to generate offline preference labels. While highly scalable, Arena Learning predominantly addresses open-domain dialogues and lacks mechanisms for domain-specific knowledge retrieval or incorporating human feedback. Consequently, it does not directly tackle real-world preference drift issues inherent in dynamic environments.

Our approach integrates and extends these methodologies not only by collecting online human preference feedback, but also explicitly gathering feedback on knowledge relevance and missing knowledge. Similarly to Arena Learning, we incorporate an LLM-based virtual judge (VJ) to filter data quality. By carefully aligning annotation methods with our training pipeline, we achieve update cycles comparable to Arena Learning's automated data flywheel, with key differences summarized in Table~\ref{tab:comparison_arena_aitl}.

\begin{table*}[t]
\centering
\small
\resizebox{\textwidth}{!}{
\begin{tabular}{lcc}
\toprule
\textbf{Aspect} & \textbf{Arena Learning} & \textbf{AITL (ours)} \\
\midrule
Feedback Source & AI-generated annotations (simulated self-play) & Human-generated annotations (real-world interaction) \\
Annotation Mode & Offline annotation & Real-time (or near-real-time) annotation \\
Update Frequency & Weekly & Weekly \\
Target Modules & Generation module only & Retrieval, Ranking, and Generation modules \\
Handling of Human Preference Drift & Indirect (no real-time human feedback) & Direct (real-time human feedback integration) \\
Verification Process & LLM-based virtual judge & LLM-based virtual judge combined with sampled human verification \\
\bottomrule
\end{tabular}}
\caption{Comparison between Arena Learning and AITL on feedback mechanisms and training pipelines.}
\label{tab:comparison_arena_aitl}
\end{table*}

\section{Method}
\label{sec:method}

\begin{figure}[t]
    \centering
    \includegraphics[width=0.98\columnwidth]{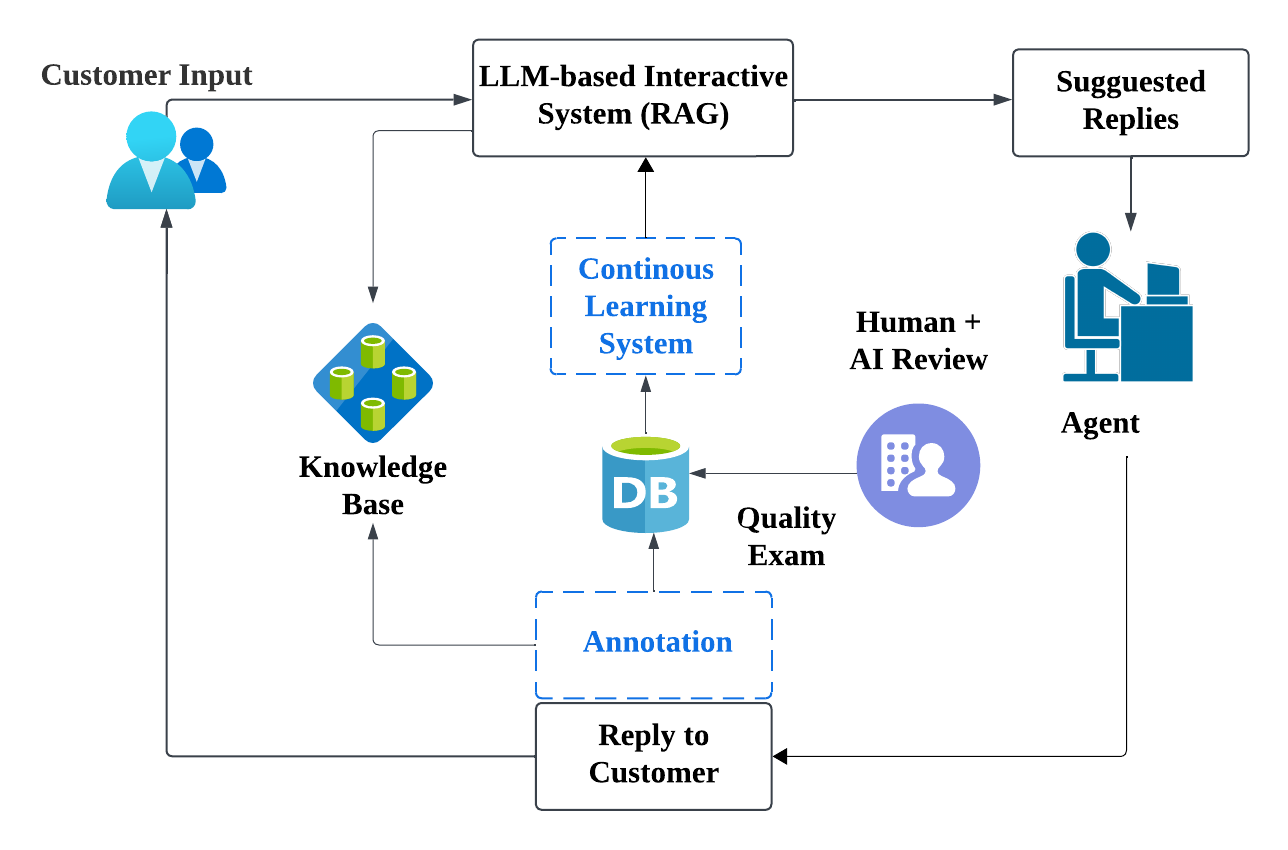} 
    \caption{Overview of the agent-in-the-loop architecture.}
    \label{fig:aitl-workflow}
\end{figure}

Figure~\ref{fig:aitl-workflow} illustrates the interactive workflow in the following key steps:
(1) \emph{Customer Input}: A customer sends a query or message.
(2) \emph{LLM-Based Interactive System}: The system retrieves relevant knowledge (Sect.~\ref{sec:unified_knowledge}) and uses an LLM to generate response candidates.
(3) \emph{Suggested Responses}: The system presents two alternative responses, potentially originating from different models.
(4) \emph{Agent Annotation}: A support agent evaluates these suggestions while serving customers, indicating their preferred response, adoption decision, critical feedback, assessment of the relevance of the knowledge used by LLM, and adding any necessary missing information (Sect.~\ref{sec:online_annotation}).
(5) \emph{Review Annotation}: Both a human expert and the LLM-based verifier review the annotation and the agent–customer interaction to flag any conflicts.(Sect.~\ref{sec:review_annotation}).
(6) \emph{Continuous Learning}: The annotations and feedback collected are reintegrated into the training pipeline for continuous model improvements (Sect.~\ref{sec:continuous_learning}).

\subsection{Unified Knowledge Base}
\label{sec:unified_knowledge}

\begin{figure}[t]
    \centering
    \includegraphics[width=0.98\columnwidth]{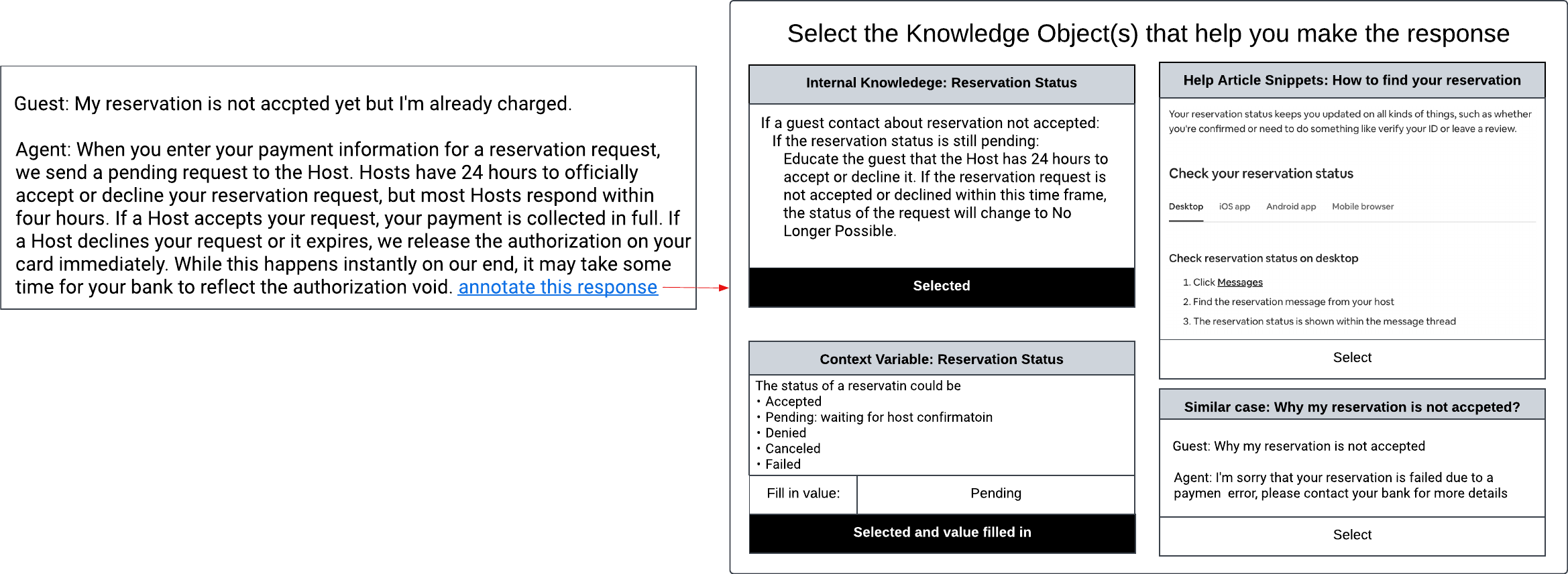} 
    \caption{Example of selecting knowledge references}
    \label{fig:ko_example}
\end{figure}

We consolidate diverse domain resources, customer guides, FAQs, internal policies, workflows, dynamic context (e.g. reservation status) and historical cases into a \emph{Unified Knowledge Base} (Figure~\ref{fig:ko_example}). The resources are enriched with detailed metadata in a centralized content management system, which facilitates the annotation and retrieval of agents in real time.

\subsection{Agent Annotation}
\begin{figure}[ht]
    \centering
    \fbox{\includegraphics[width=1.0\linewidth]{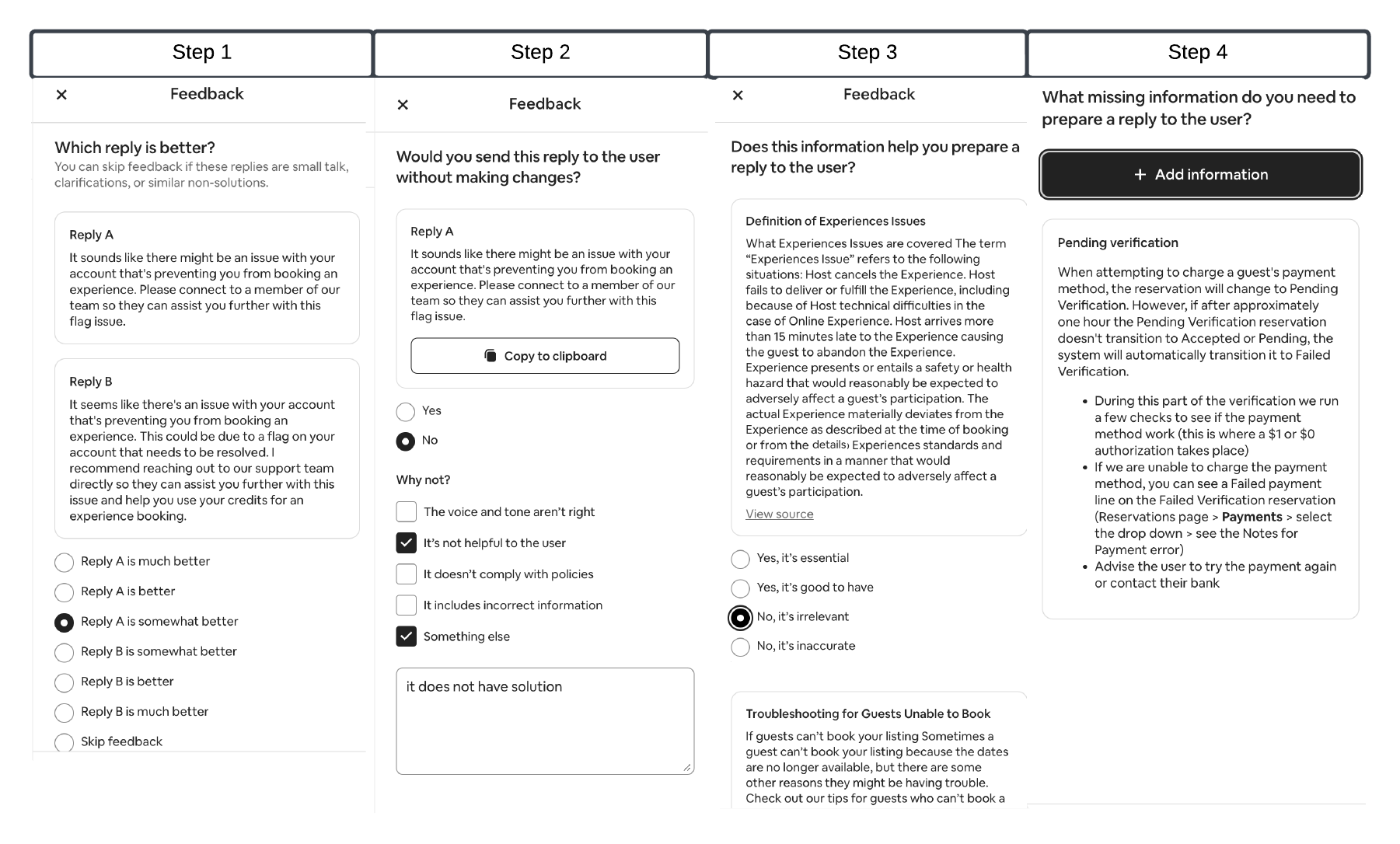}}
    \caption{Online annotation interface.}
    \label{fig:online-annotation}
\end{figure}

\label{sec:online_annotation}
Figure~\ref{fig:online-annotation} illustrates our online annotation workflow, which comprises four main steps:

\paragraph{Step 1: Pairwise Response Preference} Agents compare randomly ordered candidate responses and annotate degrees of preference as \emph{significantly better}, \emph{better}, or \emph{slightly better}. These signals inform preference learning and help to improve generation models.

\paragraph{Step 2: Rationale for Response Selection} Agents provide the adoption decision and rationales in free text (critiques), which supports the improvement of the generation model and a broader evaluation.

\paragraph{Step 3: Relevance of knowledge resources}
Agents assess and score the relevance of the knowledge resource used in the prompt, performing an additional verification in real time. These annotations directly enhance the retrieval process across diverse support topics.

\paragraph{Step 4: Missing Knowledge Identification}
During this stage, agents use a dedicated Knowledge Resources Selection Interface to flag missing information, such as policies or unrecorded best practices, that they rely on to help customers. By integrating these newly identified gaps and references back into the training pipeline, the system can improve existing retrieval recall and continuously adapt to the evolving knowledge landscape of live customer support.

\subsection{Review Annotation}
\label{sec:review_annotation}
Human and LLM-based verifiers assess the consistency between agent annotations and actual interactions, identifying common errors: preference mismatches, incorrect knowledge relevance, adoption discrepancies, and omitted knowledge (examples in Appendix~\ref{sec:appendix_annotation_errors}). The human and LLM verifier evaluations show a strong correlation (Appendix~\ref{appendix:annotation_quality_immediate_vs_delayed}).

\subsection{Continuous Learning Pipeline}
\label{sec:continuous_learning}
\begin{figure}[t]
    \centering
    \includegraphics[width=0.98\columnwidth]{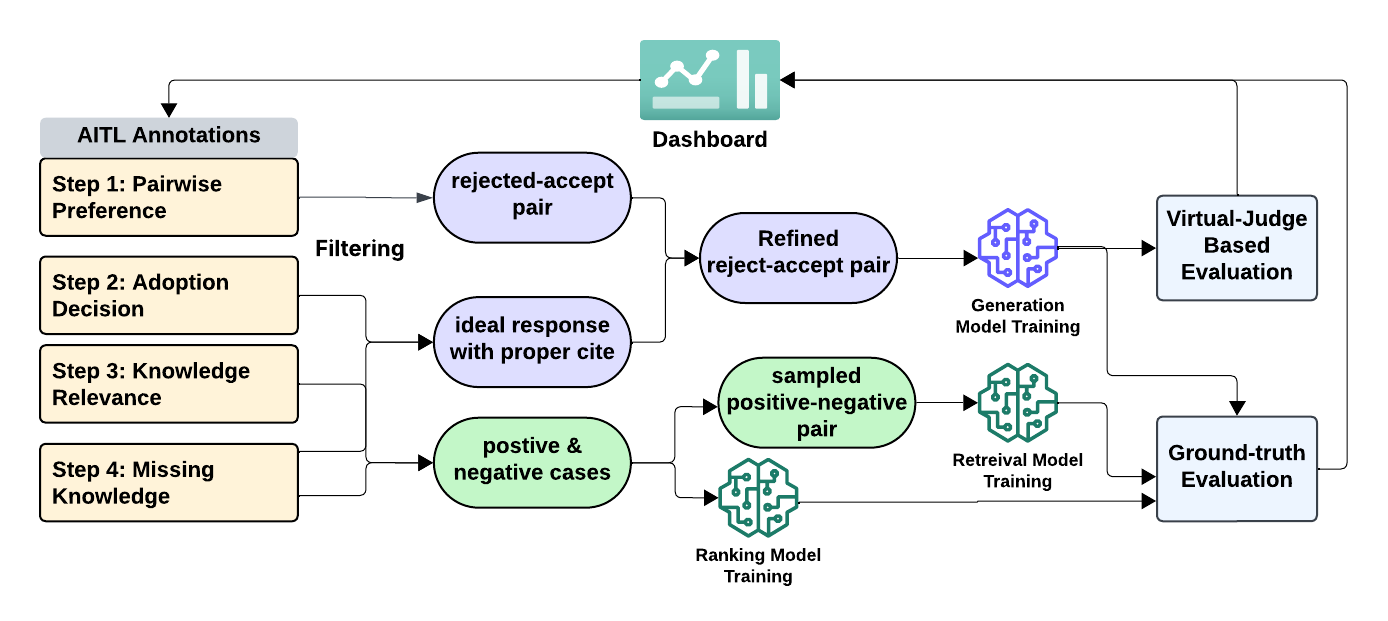}
    \caption{Data flow in continuous learning pipeline}
    \label{fig:continuous-learning}
\end{figure}

\noindent Figure~\ref{fig:continuous-learning} outlines our automated pipeline for periodic model retraining and evaluation using agent feedback. It utilizes generalized LLM offline workflow (GLOW) modules, optimizing resource usage with parameter-efficient fine-tuning (PEFT) and model partitioning (Appendix~\ref{appendix:training_experiment_efficiency}).

\paragraph{(1) Data Aggregation and Filtering}
Annotations collected from the four-step evaluation process are filtered using both rules-based and model-driven approaches. The rule-based method applies thresholds based on review scores and selects annotations that meet significant preference criteria. The model-driven approach employs an LLM-based virtual judge to filter annotations that exhibit low prompt adherence scores \citep{zheng2023judging}. These filtering strategies mitigate data inconsistencies and hallucinations, which are critical to improving model performance (Section~\ref{subsec:rag_performance}).

\paragraph{(2) Automated Model Retraining}
 Retrieval, ranking, and generation models are periodically retrained using GLOW-managed \emph{Ray} clusters, with automated resource handling and synchronization. Parameter-efficient fine-tuning (e.g., LoRA/QLoRA) optimizes GPU usage, and built-in monitoring ensures stable progress.

\paragraph{(3) Evaluation}
Retrained models go with batch inference runs on curated evaluation datasets, evaluated using both ground-truth-based evaluation and virtual judges (Appendix~\ref{appendix:helpfulness-metric-definition}) that simulate human evaluations. Performance improvements act as a \textit{proxy for annotation quality}, strengthening the cycle of enhancement.

\begin{figure}[t]
    \centering
    \includegraphics[width=0.98\columnwidth]{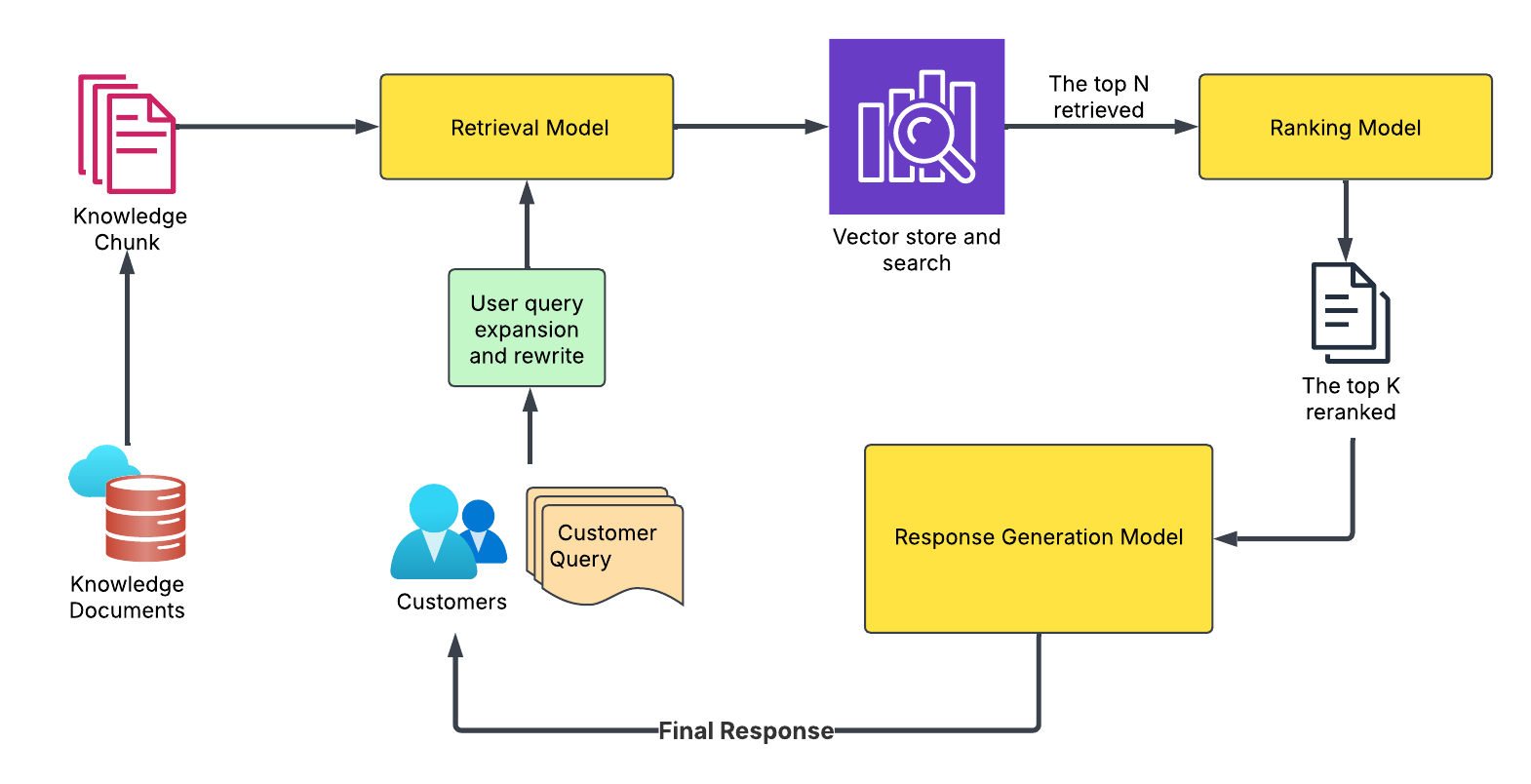} 
    \caption{RAG example on knowledge document}
    \label{fig:article_rag}
\end{figure}

\paragraph{(4) Feedback Loop}
Retrained models are deployed back into the RAG system, completing the feedback loop (Figure~\ref{fig:article_rag}).

\section{Experiments}
\label{sec:experiments}
\begin{figure}[t]
\centering
\includegraphics[width=1.0\columnwidth]{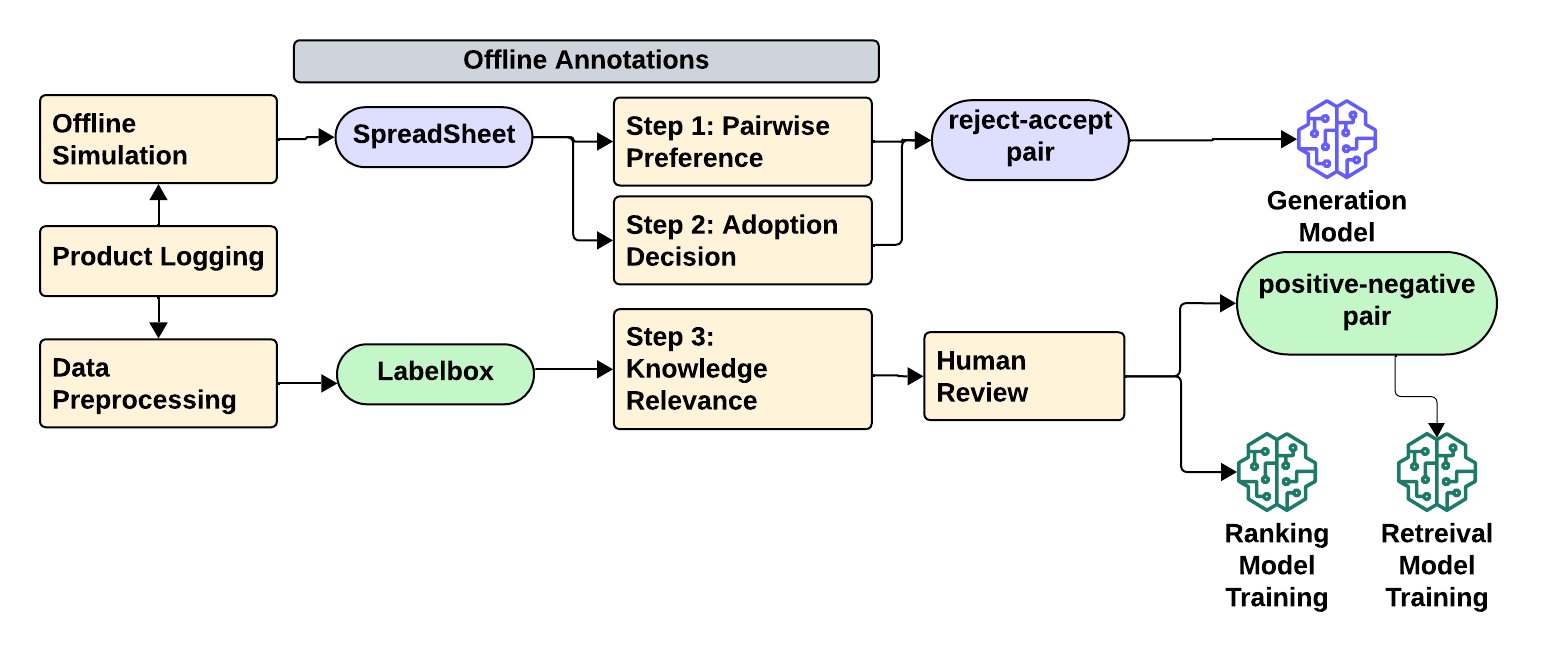}
\caption{Offline Annotation Workflow}
\label{fig:aitl-offline-annotation}
\end{figure}

Our baseline offline workflow (Figure~\ref{fig:aitl-offline-annotation}) pre-processes production logs and simulation data into preference and adoption annotations using spreadsheets, and knowledge relevance annotations via Labelbox, both validated through human review. These annotations produce rejection-accept pairs for generation tasks and positive-negative pairs for retrieval and ranking models, establishing the baseline performance prior to AITL deployment. Updating models with this offline annotation pipeline took three months.

Subsequently, we deployed the AITL annotation system into a production environment with 40 agents supporting US-based customers via an asynchronous messaging channel, where SLA requires responses within hours. Over the course of the experiment, we collected annotations from more than 5,000 customer support cases. Each agent annotated approximately 11 cases daily alongside their regular customer assistance tasks, maintaining productivity levels comparable to agents not participating in annotation tasks. The primary goal of our experiment was to compare the new AITL framework with the existing setup, evaluating its impact on annotation quality, model development cycle efficiency, and overall performance of our LLM-based customer support system.

\subsection{Evaluation Metrics}
\label{subsec:metrics}
We evaluated the effectiveness of our AITL pipeline based on annotation quality and model performance.

\paragraph{Annotation Quality.} The quality of the annotations was evaluated by averaging human experts and LLM verifiers on detecting inconsistencies between the annotations of the agents and their responses to the customers (Section~\ref{sec:review_annotation}). Reliability was measured by agreement scores, where a higher agreement indicates fewer annotation-response conflicts. The reviewers were blinded to the agent labels to reduce bias.

\paragraph{Model Performance.} We measure retrieval with \textbf{ recall @ 75} (proportion of relevant documents among the top 75 given a total of ten thousand documents) and \textbf{ precision @ 8} (relevance of the top eight ranked documents) as empirically optimal. (Appendix~\ref{appendix:retrieval_ranking_metrics}). For generation, we evaluate:

\begin{itemize} 
\item \textbf{Helpfulness}: 
\begin{enumerate}
        \item \textbf{Point-wise Helpfulness (Model-based):} Combines scores from a trained preference model and an LLM-based evaluation aligned with business criteria (Appendix~\ref{appendix:helpfulness-metric-definition}).

        \item \textbf{Pair-wise Helpfulness (Human-based):} Human annotators perform pairwise comparisons between responses, validating model-based assessments.(outlined in online annotation step 1)
\end{enumerate}
\item \textbf{Citation Correctness}: Measured as Jaccard overlap between model (\(M\)) and human (\(H\)) cited references:
\(\frac{|M \cap H|}{|M \cup H|}\).
E.g., if \(M=\{a,b\}\), \(H=\{a,c,d\}\), score is \(\frac{|\{a\}|}{|\{a,b,c,d\}|}=0.25\).
\item \textbf{Response Correctness}: Checks the factual accuracy and the adherence to the policies through the review of agents. 
\end{itemize}

\subsection{Annotation Quality Comparison}
\label{sec:anno_quality}
High-quality annotations drive a robust data flywheel, enhancing model performance and producing richer data for future cycles. Table~\ref{tab:annotation_comparison} shows higher annotation agreement rates in the online workflow compared to offline across three steps: preference, adoption, and knowledge relevance. Step~4 (missing knowledge) was not evaluated offline due to the limitations of the annotation tool in annotating all the potential missing knowledge. 

\begin{table}[h!]
  \centering
  \small
  \begin{tabular}{lcc}
    \toprule
    & \textbf{Offline} & \textbf{Online} \\
    \midrule
    \textbf{Step 1 (Preference Judgment)} & 0.635 & 0.832 \\
    \textbf{Step 2 (Adoption Judgment)}   & 0.721 & 0.775 \\
    \textbf{Step 3 (Knowledge Relevancy)} & 0.436 & 0.923 \\
    \bottomrule
  \end{tabular}
    \centering
  \small
    \caption{Agreement for offline vs.\ online setup}
      \label{tab:annotation_comparison}
\end{table}

\subsection{Impact on Model Performance}
\label{subsec:rag_performance}
We evaluated how AITL annotations and optimized fine-tuning impact retrieval-augmented generation (RAG) system performance against our baseline (Appendix~\ref{appendix:rag_baseline_details}). Models were trained using a 90\%/10\% temporal split for training and evaluation, respectively, comparing AITL annotations to offline annotations (Steps 1-3).

\paragraph{Retrieval Accuracy.}
AITL annotations significantly outperform offline annotations (Table~\ref{tab:ranker_comparison}). Precision@8 improved from 0.357 to 0.410 (+14.8\%), exceeding offline by 4.1\%. Recall@75 increased from 0.634 to 0.708 (+11.7\%), surpassing offline by 3.8\%, highlighting the benefits of AITL system.

\begin{table*}[ht]
\centering
\small
\setlength{\tabcolsep}{6pt}      
\renewcommand{\arraystretch}{1.0}
\begin{tabular}{llcc}
\toprule
\textbf{Retrieval Model} & \textbf{Ranking Model} & \textbf{Recall@75} & \textbf{Precision@8} \\
\midrule
Baseline &  Baseline & 0.634 & 0.357 \\
Offline Data Fine-tuned & Offline Data Fine-tuned & 0.670 & 0.394 \\
AITL Fine-tuned & AITL Fine-tuned & \textbf{0.708} & \textbf{0.410} \\
\bottomrule
\end{tabular}
\caption{Performance comparisons on recall@75 and precision@8 on AITL test set. The bolded entries indicate the highest scores.}
\label{tab:ranker_comparison}
\end{table*}

\paragraph{Generation Quality}
Applying ORPO \citep{hong2024orpo} with AITL fine-tuning further improved generation quality (Table~\ref{tab:rag_performance}): Helpfulness rose from 0.658 to 0.713 (+8.4\%), exceeding offline fine-tuning (0.691); Citation accuracy improved significantly from 0.097 to 0.134 (+38.1\%), surpassing offline (0.112); Response correctness increased from 0.851 to 0.882 (+3.6\%), higher than offline results (0.868). These results highlight the clear advantages of the AITL system over offline annotation pipelines (Appendix~\ref{appendix:post_training_details}).

\begin{table*}[ht]
\centering
\small
\setlength{\tabcolsep}{6pt}
\renewcommand{\arraystretch}{1.0}
\begin{tabular}{llccc}
\toprule
\textbf{Generation Model} & \textbf{Retrieval Models} & \textbf{Helpfulness} & \textbf{Citation} & \textbf{Response Correctness} \\
\midrule
Baseline & Baseline &  0.658 & 0.097 & 0.851 \\
Offline Fine-tuned (ORPO) & Offline Fine-tuned & 0.691 & 0.112 & 0.868 \\
AITL Fine-tuned (ORPO) & AITL Fine-tuned & \textbf{0.713} & \textbf{0.134} & \textbf{0.882} \\
\bottomrule
\end{tabular}
\caption{Performance of RAG models on the AITL test set. Bold entries indicate the highest scores.}
\label{tab:rag_performance}
\end{table*}    

\begin{figure}[h!]
    \centering
    \resizebox{0.48\textwidth}{!}{
        \begin{tikzpicture}
            \pie[
                text=legend,
                radius=1.5,
                explode=0.1,
                color={gray!40, blue!50, green!60},
                sum=auto,
                after number = \%
            ]{
                6.57/No Difference,
                33.32/Baseline,
                60.12/AITL Fine-tuned Models
            }
        \end{tikzpicture}
    }
    \caption{Human preference for end-to-end performance: baseline vs. AITL fine-tuned models.}
    \label{fig:pairwise_comparison_pie}
\end{figure}
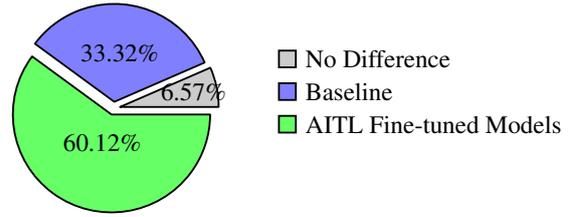

\paragraph{Human Preferences and Adoption.}
Pairwise human evaluations (Fig.~\ref{fig:pairwise_comparison_pie}) showed that 60.12\% of the fine-tuned model responses were preferred over baseline (33.32\%), and 6.57\% did not express preference. This improvement also increased the overall adoption rate by 4.5\% compared to the baseline. These findings confirm that integrating AITL annotations with ORPO not only improves objective metrics (e.g., Precision@8 and citation correctness), but also aligns with human judgments.

\section{Learnings}
\label{sec:learnings}

\subsection{Annotation Timing Ablation Study}
\begin{figure}[t]
    \centering
    \includegraphics[width=0.98\columnwidth]{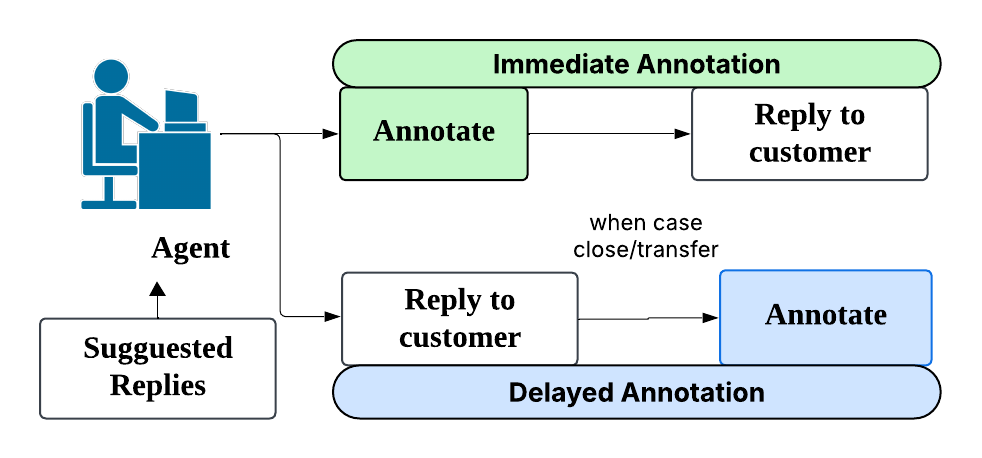}
    \caption{Immediate vs.\ Delayed Annotation Workflow}
    \label{fig:immediate_vs_delayed}
\end{figure}

A key concern is that our system design may not scale effectively to channels with stricter SLA requirements, such as live-chat. To enable annotation with higher SLA channel, we conducted a controlled experiment using the AITL tool, comparing annotations performed immediately during customer interactions (\emph{Immediate Annotation}) with those completed after interactions ended (\emph{Delayed Annotation}). This experiment involved approximately 2,000 cases under identical tooling conditions (Figure~\ref{fig:immediate_vs_delayed}).

\begin{figure}[htbp]
\centering
\hspace*{-0.8cm}
\begin{tikzpicture}

\begin{axis}[
    width=0.45\textwidth,
    height=0.35\textwidth,
    ybar,
    bar width=5pt,
    symbolic x coords={Step1,Step2,Step3,Step4},
    xtick=data,
    ymin=0, ymax=110,
    enlarge x limits=0.15,
    ylabel={Agreement (\%)},
    xlabel={Annotation Step},
    axis y line*=left,
    ymajorgrids,
    tick label style={font=\small},
    label style={font=\small},
    legend style={
        at={(0.5,1.15)},
        anchor=south,
        legend columns=3,
        font=\small
    },
]

\addplot[fill=blue!25] coordinates {
    (Step1,66.90)  
    (Step2,75.15)  
    (Step3,95.10)  
    (Step4,63.95)  
};

\addplot[fill=blue!60] coordinates {
    (Step1,67.45)  
    (Step2,75.45)  
    (Step3,94.60)  
    (Step4,76.45)  
};

\addplot[
    mark=*,
    color=red,
    thick,
    nodes near coords,
    every node near coord/.append style={
        font=\scriptsize,
        anchor=south,
        xshift=10pt
    },
    point meta=explicit symbolic,
] coordinates {
    (Step1,13.0) [1.30\,min]
    (Step2, 9.5) [0.95\,min]
    (Step3,18.7) [1.87\,min]
    (Step4,16.3) [1.63\,min]
};


\draw[semithick]
  (axis cs:Step4,85) -- (axis cs:Step4,100)
  node[midway,above,font=\small] {**}
  -- (axis cs:Step4,85);

\legend{Delayed, Immediate, Annotation Time}
\end{axis}

\begin{axis}[
    width=0.45\textwidth,
    height=0.35\textwidth,
    symbolic x coords={Step1,Step2,Step3,Step4},
    xtick=\empty,            
    ymin=0, ymax=3,          
    axis y line*=right,
    axis x line=none,
    ylabel={Annotation Time (median, min)},
    enlarge x limits=0.15,
    ylabel near ticks,
    tick label style={font=\small},
    label style={font=\small},
]
\addplot[opacity=0] coordinates {(Step1,0)};
\end{axis}

\end{tikzpicture}

\caption{Hybrid agreement (LLM + human) for immediate vs.\ delayed annotation. Red dots show median labeling time (min). ** denotes a significant difference (p<0.05).}
\label{fig:combined_annotation_agreement_time}
\end{figure}
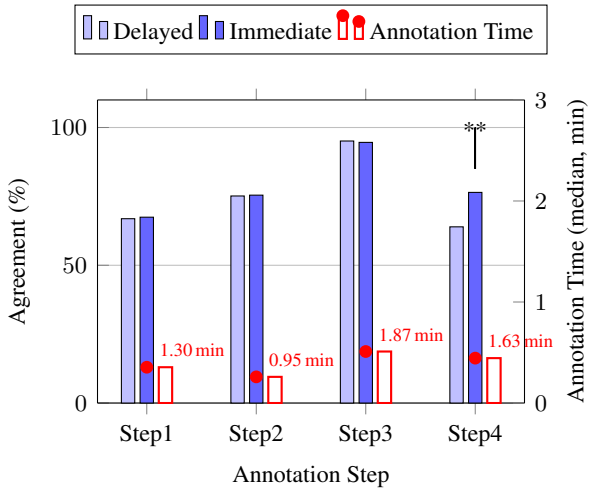

Results (Figure~\ref{fig:combined_annotation_agreement_time}) indicate that immediate annotation significantly improves annotator agreement only for the \emph{Missing-Knowledge} step (Step~4), increasing from 63.9\% to 76.5\% (+12\,pp, $p<0.05$). Conversely, differences for Preference, Adoption, and Knowledge Feedback steps (Steps~1--3) are negligible (complete scores in Appendix~\ref{appendix:annotation_quality_immediate_vs_delayed}). We therefore recommend adopting a hybrid workflow: perform immediate annotation for Missing-Knowledge when SLA allows brief delays, while delaying the remaining annotation steps after replying to customers to reliably meet stringent SLA requirements.

\subsection{Annotation Quality Ablation Study with LLM-based filter}
\label{sec:vj_ablation}

This subsection quantifies the contribution of LLM-based filter used in our data aggregation and filtering stage (Sect.~\ref{sec:continuous_learning}). The functions as a quality gate that minimizes the lack of prompt adherence and flags potential inconsistencies between the annotation of human agents and their responses to customers (Sect.~\ref{sec:review_annotation}). We do not use LLM to originate missing-knowledge labels, which require operational nuance.

\paragraph{Setup.} We evaluated two otherwise identical pipelines on the same AITL annotation batches: (i) LLM-based prompt-adherence filtering and consistency checks during data aggregation and (ii) this gate disabled. Under LLM-based filtering, 14.3\% of examples are removed prior to retriever/ranker training and 34.5\% prior to generator training.

\sisetup{
  table-number-alignment = center,
  detect-weight = true,
  detect-inline-family = math
}
\newcolumntype{Y}{S[table-format=1.3]} 

\begin{table}[t]
\centering
\footnotesize
\setlength{\tabcolsep}{3.5pt}      
\renewcommand{\arraystretch}{0.98} 
\begin{tabularx}{\columnwidth}{@{}l X Y Y@{}}
\toprule
\textbf{Model} & \textbf{Metric} & \textbf{w/ Filtering} & \textbf{w/o Filtering} \\
\midrule
\multirow{2}{*}{Retrieval}
& Recall@75            & \bfseries 0.708 & 0.670 \\
& Precision@8          & 0.402           & 0.394 \\
\midrule
\multirow{3}{*}{Generation}
& Helpfulness          & 0.703           & 0.696 \\
& Citation             & \bfseries 0.131 & 0.112 \\
& Response Correctness & 0.880           & 0.880 \\
\bottomrule
\end{tabularx}
\caption{VJ filter ablation on identical AITL batches.
The only difference is the presence/absence of VJ gating during data filtering (Sect.~\ref{sec:continuous_learning}).
Gains concentrate in retrieval recall and citation accuracy.}
\label{tab:vj_ablation}
\end{table}

\paragraph{Findings.}
Table~\ref{tab:vj_ablation} shows that LLM-based filtering yields consistent gains in Recall@75 (+3.8\,pp vs.\ no-filtering on the same batch) and Citation (+1.9\,pp absolute), while Precision@8, Helpfulness, and Response Correctness remain statistically unchanged. This pattern aligns with the LLM’s role as a noise gate: by down-weighting low–prompt-adherence or hallucination-prone supervision during data filtering, it primarily improves retrieval and citation grounding without perturbing other signals.

\paragraph{Interpretation.}
As detailed in Appendix~\ref{appendix:annotation_quality_immediate_vs_delayed}, LLM and human evaluations are strongly correlated ($r\!>\!0.90$), enabling a cost-effective hybrid reliability score that combines both sources. A notable exception is Step~4 (\emph{Missing-Knowledge}), which benefits uniquely from immediate human annotation (+12\,pp agreement; Fig.~\ref{fig:combined_annotation_agreement_time}). Accordingly, we position the LLM-based filtering as a validator rather than a generator for this channel, yielding a scalable quality filter while preserving human oversight where domain nuance is essential.

\subsection{Pairwise Preference Data Effectiveness.}
Fine-tuning with plus-level preference data (better or significantly better) boosts helpfulness, but lowers the correctness of the citation. Adding an agent adoption as a data filter restores the correctness of the citation to $11.4\%$ and retains a gain of $3.5\%$ in helpfulness, achieving a better balance. 

\begin{table}[h!]
\centering
\small
\begin{tabular}{lcc}
\hline
\textbf{Fine-tune Strategy} & \textbf{Helpfulness} & \textbf{Citation} \\
\hline
Baseline            & 0.694          & 0.123 \\
Plus Pref           & \textbf{0.766} & 0.109 \\
Plus Pref + Adopted & 0.718          & \textbf{0.137} \\
\hline
\end{tabular}
\caption{Performance comparison of generation models by different preference dataset}
\end{table}

\subsection{Continuous Training Strategies.}
Retraining the models using a mix of historical and new annotations on the previous checkpoint improves adaptability and robustness, increasing precision @ 8 by 8\% in historical data and 4\% in recent data compared to training only on new data. (Appendix~\ref{appendix:post_training_details}) Periodically integrating fresh feedback with diverse historical datasets mitigates overfitting and improves retrieval, ranking, and generation performance. Similar benefits are observed with offline annotation approaches.

\subsection{Cross-Model Generalization Study}
\label{sec:cross_model_ablation}
We replicate AITL on Qwen2.5-32B and the Llama-3 family (3B, 8B, 70B) using the same data split and evaluator as the main study. As shown in Table~\ref{tab:cross_model_all}, AITL delivers consistent gains on smaller and medium models. At 70B, where SFT baselines are already strong, AITL shows mixed helpfulness effects but improved/stable citation across variants, suggesting interaction with prior objectives rather than a lack of transfer; overall, these results indicate the AITL data flywheel generalizes across architectures and scales.

\begin{table}[t]
\centering
\small
\begin{tabular}{lccc}
\toprule
\textbf{Model} & \textbf{AITL} & \textbf{Helpfulness} & \textbf{Citation} \\
\midrule
Qwen2.5-32B-Instruct & No  & 0.6718 & 0.1040 \\
Qwen2.5-32B-Instruct & Yes & \textbf{0.6830} & 0.1040 \\
Llama-3.2-3B-Instruct & No  & 0.3731 & 0.0569 \\
Llama-3.2-3B-Instruct & Yes & \textbf{0.6362} & \textbf{0.0606} \\
Llama-3.1-8B-Instruct & No  & 0.3787 & 0.0967 \\
Llama-3.1-8B-Instruct & Yes & \textbf{0.6056} & \textbf{0.1136} \\
Llama-3.3-70B-Instruct & No  & 0.6322 & 0.1048 \\
Llama-3.3-70B-Instruct & Yes & \textbf{0.6438} & \textbf{0.1224} \\
\bottomrule
\end{tabular}
\caption{Cross-model results with the same AITL split and evaluator.}
\label{tab:cross_model_all}
\end{table}

\section{Conclusion and Future Work}
\label{sec:conclusion_and_future_work}
We introduced \emph{Agent-in-the-Loop} (AITL), an real-time(near real-time) data flywheel that turns routine customer support operations into continuously improving supervision for retrieval, ranking, and generation. By capturing four signal types including pairwise response preferences, agent adoption and rationales, knowledge relevance, and missing knowledge, AITL closes the gap between evaluation and production reality. Our pilot with U.S.-based agents shows consistent gains in retrieval (Recall@75, Precision@8), generation helpfulness, citation correctness, and agent adoption, while shrinking update cadence from months to weeks. 

\noindent Building on these results, we outline three directions for future work:
\begin{itemize}
  \item \textbf{Scaling optional agent feedback.} Replace heavy labels with lightweight micro-annotations (default ``skip''), use active sampling for high-uncertainty or disagreement cases, and correct selection bias via inverse-propensity weighting and post-stratification.
  \item \textbf{Product-embedded AITL for efficiency.} Integrate AITL into agent-facing tools; evaluate with a productivity bundle (e.g., CSAT, time-to-resolution, adoption rate, human-edit distance); and study cognitive load, trust calibration, and skill formation across novice and expert agents.
  \item \textbf{Toward fuller automation.} Leverage simulation and judge-based validation to automate dataset curation and preference labeling where appropriate, while preserving human oversight for safety, policy adherence, and domain nuance.
\end{itemize}

\section*{Limitation}
Although AITL offers clear advantages, there are three key limitations. 

First, prolonged use of real-time annotations could lead to increased agent workload and potential annotation fatigue. To mitigate this risk, future implementations could consider strategies such as rotating annotation responsibilities among agents, adaptive workload management, and periodic breaks from annotation duties. In addition, targeted training and incentive programs can further support annotation quality over time.

Second, our study exclusively focused on English-language customer support. This leaves open questions regarding the effectiveness and applicability of AITL in multilingual or culturally diverse support contexts, which should be investigated in future research.

Finally, the relatively short duration of this study constrains our understanding of how annotation practices might evolve over extended periods and, crucially, how effectively they scale when applied to larger groups of agents.

\section*{Acknowledgments}
We are deeply grateful to the many colleagues who contributed to the AITL project. This work was made possible by the GLOW platform, which enabled reproducible workflows, streamlined model development, and rapid end-to-end iterations.

We extend our sincere thanks to the leadership of Airbnb Customer Support for their sponsorship and guidance. We especially thank Hossein Shams, Chloe Zhao, Gen Wang, Chandraprakash Loonker, Mengchen Liang, Jeremy Wang, Jingwen Qiang and Ying Tan for their engineering contributions.

We thank Tony Donisch and Chris Robinson for project ideation and support. We are also grateful to our colleagues in Design including Eric Fensterheim, Stacey Kennelly Nester, whose creativity and insights helped shape the project. In Platform Management, we thank Omar Siddiqui, Eliav Kahan, and Jorge Poblacion for their direction and support. Most importantly, this project could not have been completed without the partnership of our agent support collaborators; we acknowledge Lindsey Oben, David Amador, Isabel Arboleda, Chris Enzaldo, and the CS Labs team for their indispensable support. Finally, we thank the anonymous reviewers for their constructive feedback, which helped us substantially improve this paper.

\bibliography{aitl/custom}

\appendix
\section{Annotation Error Types and LLM Verifier Prompt}
\label{sec:appendix_annotation_errors}
We identify common annotation error types, each illustrated with concise examples:
\begin{itemize}
\item \textbf{Preference Mismatch}: Occurs when an agent explicitly prefers response A but ultimately selects response B.
\item \textbf{Adoption Mismatch}: Occurs when an agent rejects a response due to specific formatting issues but subsequently uses the same formatting in their final reply.
\item \textbf{Incorrect Knowledge Annotation}: Occurs when the agent provides guidance based on certain policies (e.g., advising customer cancellation) but inaccurately labels relevant policy documents as irrelevant.
\item \textbf{Omitted Missing Knowledge}: Occurs when the agent mentions essential details (e.g., refund amount) to the customer but fails to annotate this critical contextual information.
\end{itemize}

Below is a refined example prompt template utilized by the LLM-based verifier for missing knowledge object verification:

\begin{promptbox}
\textbf{Task:} Evaluate whether the provided knowledge objects annotated by agents sufficiently address the customer's issue described in the conversation. \\

Follow these instructions step by step: \\
 - Read and understand the customer's issue from the provided conversation and contextual information. \\
 - Carefully review the agent's annotated knowledge objects and their response to the customer. \\
 - Determine if the annotated knowledge items adequately and directly resolve or address the customer's issue. \\

Provide your evaluation strictly in the following JSON format:

\{
  "missing\_agreement\_reason": "Provide a concise yet specific reason why the knowledge objects are sufficient or insufficient.",
  "missing\_agreement": 1 or 0  // 1 if knowledge provided is insufficient, 0 if sufficient
\}

\textbf{Conversation:} \\
\texttt{[Customer]}: what customer said\\ 
\texttt{[Agent]}: what agent said \\
\texttt{[Customer]}: what customer said \\

\textbf{Agent's Response to Customer:} \\
agent's final response provided here \\

\textbf{Provided Knowledge Objects Annotated by Agents:} \\
1. Knowledge item 1 here \\
2. Knowledge item 2 here \\
\end{promptbox}

\section{Helpfulness Metric Definition}
\label{appendix:helpfulness-metric-definition}
This section will describe the training process for helpfulness models.

We define \emph{point-wise helpfulness score} to measure the likelihood that internal domain experts (e.g. customer support agents) would prefer a particular response. This score is generated through an ensemble method that combines two distinct approaches:

1. Reference Evaluator Model: We employ an internally developed evaluation model based on Mistral-7B, fine-tuned using human-annotated preference data via a pairwise loss objective. This model serves as our reference evaluator, achieving an approximate agreement rate of 80\% compared to the ground-truth annotations provided by the expert groups.

2. GPT-4 Prompt-based Evaluations: We perform multiple GPT-4 evaluations using carefully designed prompts that embody internal helpfulness criteria. Each evaluation produces a binary indicator (helpful or not), and we sum these indicators across seven distinct prompts to yield an integer score ranging from 0 to 7. By averaging these indicators, we obtain a final continuous helpfulness score. This prompt-based mechanism eliminates the need for ground-truth labels during inference. Evaluation against expert-labeled ground truth confirms an agreement rate of approximately 80\%, validating the reliability of this approach.

\section{Retrieval and Ranking Metrics}
\label{appendix:retrieval_ranking_metrics}
We empirically selected Recall@75 and Precision@8 as our primary retrieval and ranking metrics, guided by operational constraints and experimental insights. Recall@75 evaluates the effectiveness of our retrieval system in identifying relevant articles within the top 75 retrieved candidates. Empirical analysis showed that the recall increased consistently with more retrieved candidates, reaching approximately 70\%-80\% at topN=75. Beyond 75 candidates, recall improvements plateaued, indicating limited benefits from retrieving additional documents.

Precision@8 was chosen based on practical limitations, as production environments restrict input context lengths for downstream generation models. Our experiments indicated that reranker performance notably surpassed retrieval-only methods from topN=5 onward, with topN=8 providing the optimal trade-off between performance enhancement and operational feasibility. Increasing the number of retrieved snippets beyond 8 led to reduced helpfulness scores due to excessive information dilution. Thus, Precision@8 effectively balances input quality for response generation with practical system constraints.

\section{Prior AITL Baseline Details}
\label{appendix:rag_baseline_details}
This appendix details the baseline Retrieval-Augmented Generation (RAG) system, outlining the three core models: generation, retrieval, and ranking.

\textbf{Generation Model}
Our generation component utilizes an 8$\times$7B Mistral Mixture-of-Experts (MoE) model \citep{jiang2024mixtral}. This model is initially fine-tuned via supervised fine-tuning (SFT) \citep{ouyang2022traininglanguagemodelsfollow} on offline human-agent annotations. Subsequently, we apply Odds Ratio Preference Optimization (ORPO) \citep{hong2024orpo} to align the model's generation with agent-approved preferences.

\textbf{Retrieval Model} Retrieval is performed by fine-tuning a pre-trained Zeta-Alpha-E5-Mistral 7b embedding model \citep{camara2024zetaalpha}. This fine-tuned model transforms article chunks into 1024-dimensional vector embeddings. When a user query is received, it is encoded, and the top-$N$ relevant article chunks are retrieved based on cosine similarity.

\textbf{Ranking Model} After retrieval, a ranking model refines the top-$N$ retrieved article chunks to identify the most relevant and helpful content. This ranking is accomplished using an in-house fine-tuned FLAN-T5 model\citep{chung2022scalinginstructionfinetunedlanguagemodels}, adapted for a ranking task. The model takes the user query and each retrieved article chunk's details as input, and outputs a relevance score, which is then used to rank the chunks.

\section{AITL Training Details}
\label{appendix:post_training_details}
This appendix details the post-training processes and parameters for each component of our Retrieval-Augmented Generation (RAG) baseline: the Generation Model, Retrieval Model, and Ranking Model.

\subsection{Generation Model}
Fine-tuning commenced from the baseline checkpoint of the 8$\times$7B Mistral Mixture-of-Experts (MoE) model. The model was optimized via Odds Ratio Preference Optimization (ORPO), using a balanced dataset comprising equal parts newly annotated examples and previously collected training data. Training was conducted for a total of 3 epochs with a batch size of 64, employing an initial learning rate of $2 \times 10^{-5}$, which linearly decayed after a warm-up period covering 5\% of total training steps. The fine-tuning process utilized 4 NVIDIA A100 GPUs, each equipped with 80GB of memory. After training, the model underwent 4-bit Quantized Low-Rank Adaptation (QLoRA) to substantially decrease computational overhead and memory consumption during inference. Subsequently, the LoRA weights were merged back into the original model weights, ensuring efficient deployment. Parameter selection and alignment loss criteria were informed by comparisons made during previous model iterations.

\begin{itemize}
    \item \textbf{Training:} 
    \begin{itemize}
        \item \emph{ORPO Phase:} Incorporates preference signals through pairwise feedback. The model learns to prioritize more 'helpful' or relevant outputs based on labeled data.
    \end{itemize}

    \item \textbf{Observations:}
    \begin{itemize}
        \item ORPO fine-tuning often yields higher coherence and factual correctness than SFT alone.
        \item We observe slight divergences between purely offline vs. online-labeled preference data through AITL, motivating continuous updates.
    \end{itemize}
\end{itemize}

\subsection{Retrieval Model}
\label{appendix:retrieval_model}

Following the generation model's optimization, we proceed to detail the training process for the retrieval model of our RAG system.

\begin{itemize}
    \item \textbf{Training:} 
    \begin{itemize}
        \item \emph{Fine-tuning Methodology:} The Zeta-Alpha-E5-Mistral 7b model underwent fine-tuning employing MultipleNegativesRankingLoss. Positive training samples were comprised of article segments marked as "RELEVANT" or "HELPFUL," supplemented by agent-generated content. Negative samples consisted of both "NOT RELEVANT" or "NOT HELPFUL" labeled chunks (hard negatives) and randomly sampled chunks from the same batch (easy negatives). 
        \item \emph{Training Setup:} Training parameters included a learning rate of $2 \times 10^{-5}$ and a weight decay of $2 \times 10^{-6}$. The training was conducted for $1$ epoch, with a training batch size of $1$. The model was executed on an A100 GPU cluster consisting of $4$ GPUs. 
    \end{itemize}

    \item \textbf{Observations:}
    \begin{itemize}
        \item The retrieval model's accuracy benefited from increased training dataset size and demonstrated resilience to data noise. Notably, optimal performance was observed when the model was trained on a combination of high-confidence (annotator and reviewer alignment) and low-confidence (annotator and reviewer disagreement) datasets.
        \item For RAG in live support conversations, the retrieval models fine-tuned using AITL data exhibited substantially superior performance compared to the baseline, and those fine-tuned using offline datasets.
    \end{itemize}
\end{itemize}

\subsection{Ranking Model}
\label{appendix:ranking_model}

With the retrieval model established, we now turn our attention to the ranking model, which refines the retrieved results.

\begin{itemize}
    \item \textbf{Training:} 
    \begin{itemize}
        \item \emph{SFT:} Training data consisted of positive and negative examples based on agent annotations. Positive examples were constructed from article chunks labeled "RELEVANT" or "HELPFUL," along with agent-generated content. Negative examples comprised chunks labeled "NOT RELEVANT" or "NOT HELPFUL." Each training instance was formatted as \textbf{<prompt + user query + article chunk details>} as input, with \textbf{<relevant>} as the target for positive examples and \textbf{<not\_relevant>} as the target for negative examples. During inference, the relevance score is determined by the probability of the output being "relevant".
        \item \emph{Training Setup:} The model was trained for 3 epochs on a single A10 GPU, using a learning rate of $1e-5$, a batch size of 16, and gradient accumulation steps of 64.
    \end{itemize}

    \item \textbf{Observations:}
    \begin{itemize}
        \item The ranking model's performance was optimized when trained on high-confidence annotated datasets, with a balanced 1:2 ratio of historical and newly generated AITL data.
        \item Utilizing AITL data for fine-tuning ranking models yielded significantly better performance in live support RAG scenarios compared to offline datasets, and also surpassed the performance of models trained only on historical data.
    \end{itemize}
\end{itemize}

\section{Training Experiment Efficiency}
\label{appendix:training_experiment_efficiency}
To improve the efficiency of offline experiments, we introduced a framework for the generalized offline LLM workflow (GLOW) based on reusable and parameterized fine-tuning, batch scoring and evaluation components. The two main areas of focus for GLOW that benefit this research study include infra-aware LLM developments and integration with template end-to-end workflow.

\paragraph{Infra-aware LLM Developments}
LLM computations are very sensitive to the underlying infrastructure, and there are multiple hyperparameters that can drastically affect the computation capacity:
\begin{itemize}
    \item Compute optimization: whether to apply lower precision training, Parameter-Efficient Fine Tuning (PEFT) adaptors and model partitioning strategies like DeepSpeed or FSDP
    \item Input dataset size, batch size and context length
    \item Model capacity and architecture
\end{itemize}
A typical LLM offline task that has a specific set of hyperparams must be deployed to a matching infrastructure to avoid failed tasks due to insufficient GPU VRAM, or overprovisioning that led to low GPU utilization rate. To solve this problem, GLOW in its configuration is integrated with the Ray Cluster spec section that combines the LLM task with ephemeral compute cluster provisioned on-demand. This setup not only improves the offline experiment task stability, but also guarantees reproducible results across multiple experiment runs.

\paragraph{Templated end-to-end Workflow}
GLOW offers reusable offline workflow building components for end users to customize their developments, and this unified API improves production readiness while largely reducing prototype development cycles to production.

\section{Immediate and Delayed Comparison}
\label{appendix:annotation_quality_immediate_vs_delayed}
In practice we average the human and LLM scores to form a hybrid metric.  Because the two sources correlate strongly ($r\!>\!0.90$; Table~\ref{tab:llm_human_sig}), this composite score inherits human‐level reliability while remaining inexpensive to scale.  All subsequent analyses, including the immediate vs.\ deferred comparison in Figure~\ref{fig:combined_annotation_agreement_time}, are therefore reported on this hybrid metric.

\begin{table}[t]
\centering
\small
\setlength{\tabcolsep}{5pt}  
\begin{tabular}{lcccc}
\toprule
\multirow{2}{*}{\textbf{Annotation Step}} &
\multicolumn{2}{c}{\textbf{Immediate (\%)}} &
\multicolumn{2}{c}{\textbf{Delayed (\%)}} \\
\cmidrule(lr){2-3}\cmidrule(lr){4-5}
 & \textbf{LLM} & \textbf{Human} & \textbf{LLM} & \textbf{Human} \\
\midrule
Step 1 & 62.6 & 72.3 & 61.0 & 72.8 \\
Step 2 & 76.7 & 74.2 & 75.1 & 75.2 \\
Step 3 & 91.7 & 97.5 & 91.7 & 98.5 \\
Step 4 & 76.1 & 76.8 & 57.6\textsuperscript{**} & 70.3\textsuperscript{**} \\
\bottomrule
\end{tabular}
\caption{Annotation accuracy for LLM and human raters.  
Significance marker \textsuperscript{**}: difference between immediate and delayed is significant ($p<0.05$ for humans; $\Delta>5$ pp for LLM).}
\label{tab:llm_human_sig}
\end{table}

\begin{table}[htbp]
\centering
\footnotesize                     
\setlength\tabcolsep{3.5pt}       
\begin{minipage}{0.48\textwidth}
\centering
\begin{tabular}{lrrrrr}
\toprule
\textbf{Step} & \textbf{Q1} & \textbf{Median} & \textbf{Q3} & \textbf{Mean} & \textbf{Trim.\ Mean} \\
\midrule
Step 1 & 0.583 & \textbf{1.30} & 2.967 & 4.017 & 2.174 \\
Step 2 & 0.483 & \textbf{0.95} & 2.733 & 3.711 & 2.065 \\
Step 3 & 1.317 & \textbf{1.87} & 3.867 & 4.647 & 3.332 \\
Step 4 & 0.721 & \textbf{1.63} & 3.317 & 3.409 & 2.430 \\
\bottomrule
\end{tabular}
\caption{Annotation-time statistics per step (minutes).}
\label{tab:annotation_time_stats}
\end{minipage}
\end{table}

\vspace{0.5em}

Table~\ref{tab:annotation_time_stats} shows a pronounced right skew: Means and even 10 \% trimmed means sit well above the medians of 1–2 min. Hence, we quote the median as the best indicator of agent annotation time effort, with the trimmed mean offered as a reference.

\end{document}